\documentclass[lettersize,journal]{IEEEtran}
\usepackage{amsmath,amsfonts}
\usepackage{array}
\usepackage[caption=false,font=normalsize,labelfont=sf,textfont=sf]{subfig}
\usepackage{textcomp}
\usepackage{stfloats}
\usepackage{url}
\usepackage{verbatim}
\usepackage{graphicx}
\usepackage{cite}
\usepackage[linesnumbered,ruled,vlined]{algorithm2e}
\usepackage{booktabs}
\usepackage{multirow}
\usepackage{hyperref}
\begin{document}

\title{BP-TTA: Balanced and Prototype-Guided Test-Time Adaptation  in Dynamic Scenarios}

\author{Shaoyang Huang, Yashi Zhu, Yichen Yu, Lei Zhang, Zhang Yi, \IEEEmembership{Fellow, IEEE}, Tao He 
\thanks{This work was supported by the Fundamental and Interdisciplinary Disciplines Breakthrough Plan of the Ministry of Education of China under Grant No. JYB2025XDXM101 and the Natural Science Foundation of Sichuan Province under Grant No. 2026NSFSC1488. (Corresponding author: Tao He, email: tao\_he@scu.edu.cn).}
\thanks{Shaoyang Huang, Yashi Zhu, Yichen Yu, and Zhang Yi are with the College of Computer Science, Sichuan University, Chengdu, 610065, China. Tao He and Lei Zhang are with the School of Artificial Intelligence, Sichuan University, Chengdu, 610065, China.}
\thanks{Shaoyang Huang and Yashi Zhu are co-first authors and contributed equally to this study.}
}

\markboth{Journal of \LaTeX\ Class Files,~Vol.~14, No.~8, August~2021}%
{Shell \MakeLowercase{\textit{et al.}}: A Sample Article Using IEEEtran.cls for IEEE Journals}


\maketitle

\begin{abstract}
Test-Time Adaptation (TTA) enables models trained on a source domain to adapt online to unlabeled test data under distribution shifts. While recent TTA methods have moved beyond static settings and begun to consider continual domain shifts, they primarily address distribution drift and fail to account for class imbalance in dynamic scenarios. In real-world test-time streams, class imbalance and continual domain shifts often occur at the same time and interact with each other. In this paper, we propose a novel Balanced and Prototype-Guided Test-Time Adaptation (BP-TTA) method, which combines batch-balanced sampling with prototype-guided adaptation to handle the class imbalance and continual domain shift problems. BP-TTA constructs balanced adaptation batches by integrating current samples with high-confidence historical instances, effectively mitigating bias toward dominant classes and stabilizing online updates. Meanwhile, BP-TTA maintains evolving class prototypes during inference and leverages prototype similarity as a constraint for model adaptation, thereby improving the reliability of pseudo-labels and enhancing the stability of online updates under persistent domain shifts. Extensive experiments demonstrate that BP-TTA consistently outperforms state-of-the-art TTA methods in dynamic test-time streaming settings.
\end{abstract}


\section{Introduction}
\IEEEPARstart{D}{eep} neural networks have achieved remarkable progress in computer vision and related fields in recent years \cite{he2016deep,krizhevsky2012imagenet}. However, their success typically relies on the assumption that training and test data follow the same distribution, which rarely holds in real-world applications. In practice, factors such as illumination changes, weather variations, sensor noise, and differences in imaging devices often lead to distribution shifts. When faced with such shifts at inference time, model performance can degrade substantially, sometimes resulting in catastrophic failures \cite{han2024model}. Since it is infeasible to train models on all possible data distributions, this reliance on the training distribution limits their adaptability. 

To mitigate distribution shifts, prior research has extensively explored Unsupervised Domain Adaptation (UDA), which addresses domain gaps by jointly training on labeled source data and unlabeled target data \cite{udp1, udp2, udp3, udp4}. However, it requires access to both source and target data during training. In practice, target data is often unavailable beforehand, distribution shifts are unpredictable, and access to source data may also be restricted due to privacy or computational constraints.
In response to this challenge, Test-Time Adaptation (TTA) has recently attracted growing attention in scenarios involving domain shifts caused by image corruptions \cite{dobler2023robust,karmanov2024efficient}. In this paradigm, a pre-trained source model, without access to its training data, is adapted to unlabeled test data through online training. 

Early TTA studies \cite{tent,bn1,adacontrast} primarily focused on static adaptation settings. These methods assumed that all test samples were independently drawn from a single fixed target domain, implying that each test batch was independently and identically distributed (i.i.d.). Under this assumption, the model can gradually adapt to a stationary target distribution by minimizing entropy or updating normalization statistics. However, such an assumption is often unrealistic in practice, as real-world test data rarely follow a fixed distribution and instead exhibit complex and evolving patterns over time. 
To address this limitation, later works \cite{cotta,eata,dss} proposed Continual Test-Time Adaptation (CTTA), which considers a more realistic dynamic adaptation scenario where the test distribution continuously shifts over time under changing environments. Furthermore, \cite{rotta} introduced a more practical setting that not only follows the continual adaptation paradigm but also accounts for the fact that test data usually arrive as streaming sequences with strong temporal correlations. This often leads to severe class imbalance within short time windows. Termed Practical Test-Time Adaptation (PTTA), this setting presents two critical challenges: class imbalance and continuous domain shifts.


First, class imbalance occurs when some classes appear frequently within a short period while others are underrepresented or temporarily absent. This situation often comes from temporal correlation in real-world streaming data. For example, in autonomous driving, vehicles on highways predominantly observe cars, whereas in residential areas they mostly encounter pedestrians or non-motorized vehicles. Such temporally localized distributions lead to biased model updates because the model repeatedly learns from dominant classes \cite{shwartz2023simplifying}. As a result, gradient updates are dominated by majority classes, causing the model to overfit to short-term patterns while neglecting minority classes, ultimately degrading overall generalization performance.

Second, continuous domain shift refers to gradual changes in the test distribution as environmental conditions evolve over time. For instance, in autonomous driving, lighting changes when vehicles move from sunny roads to rainy streets or tunnels, causing shifts in input features over time \cite{shift2022}. Under continual domain shift settings, TTA methods that rely on pseudo-labels or entropy minimization tend to be unstable and highly sensitive to noisy signals. As the distribution drifts, pseudo-labels may degrade, producing noisy supervision that misguides gradient updates and reduces model reliability. 

\begin{figure*}[t]
    \centering
    \subfloat[\label{fig:BBS}]{
        \includegraphics[height=4.8cm]{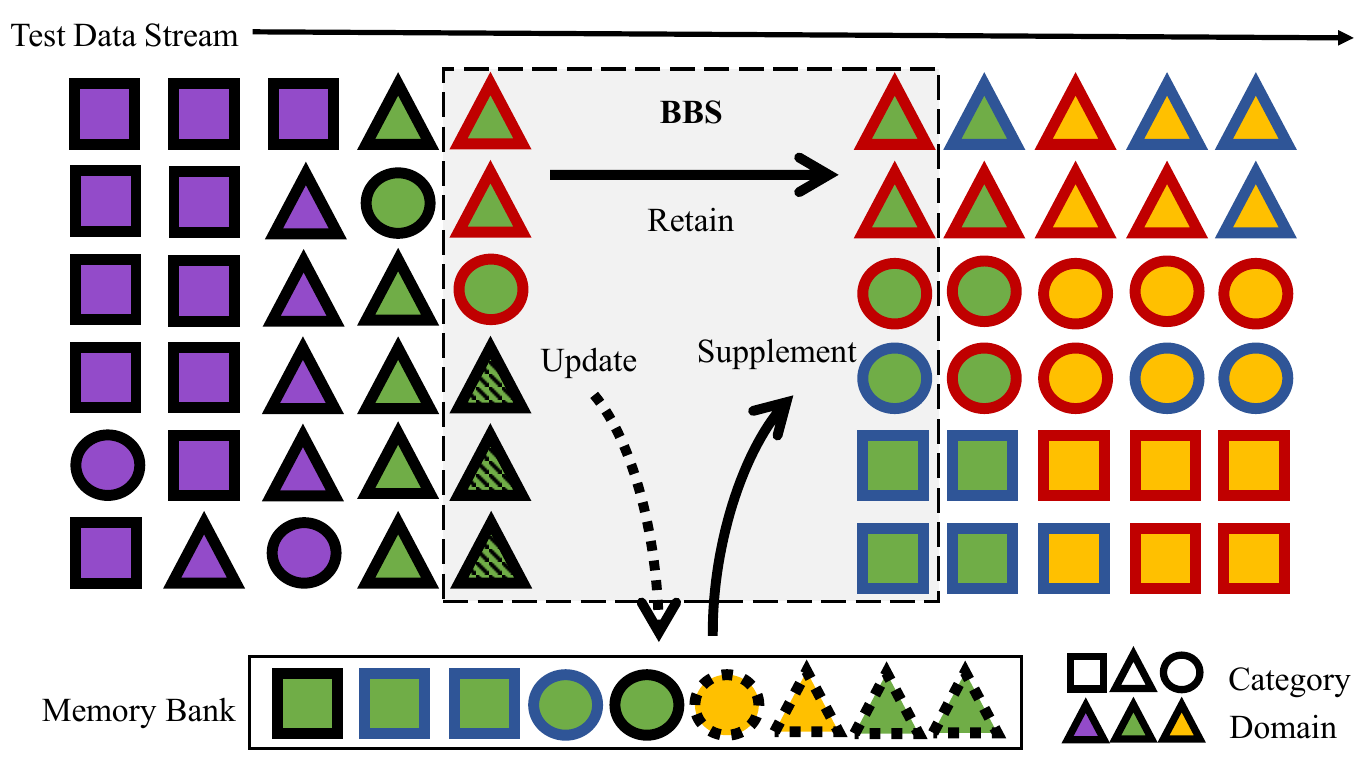}
    }
    \hfill
    \subfloat[\label{fig:CPGA}]{
        \includegraphics[height=4.8cm]{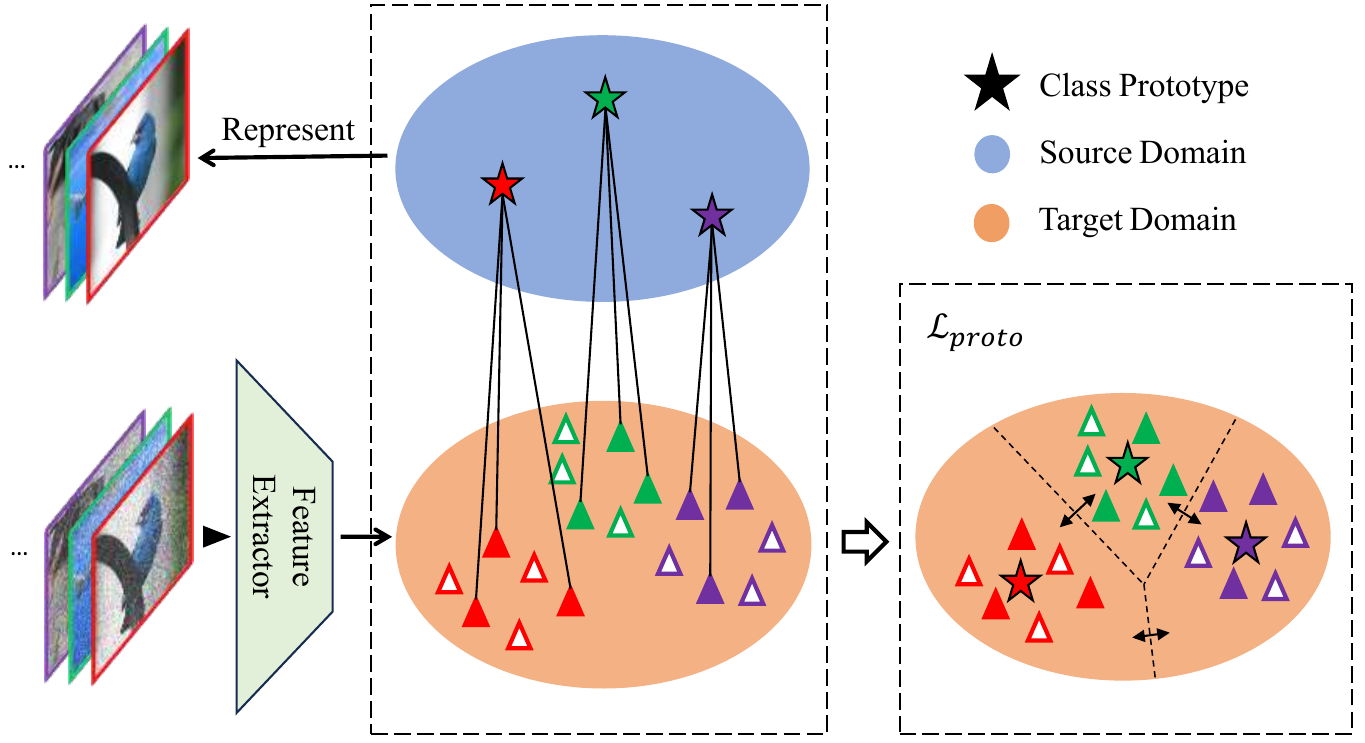}
    }
    \caption{Visualization of our two key components. (a) BBS selects high-quality samples from majority classes and supplements minority-class samples from the memory bank to construct a balanced batch for adaptation. The memory bank is selectively updated with new samples from the current batch.
    (b) CPGA first updates class prototypes using high-confidence samples's features (solid triangles), then computes the prototype loss to guide model adaptation, promoting intra-class clustering and inter-class separation.}
\end{figure*}

To jointly address class imbalance and continuous domain shift, we propose a Balanced and Prototype-Guided Test-Time Adaptation (BP-TTA) method, which combines batch-balanced sampling with prototype-guided adaptation. The following are our main contributions:
\begin{itemize}
    \item We propose Batch-Balanced Sampling (BBS) to address class imbalance, which constructs a class-balanced memory bank from realistic test streams and dynamically reconstructs the current batch according to the class distribution of incoming samples, effectively mitigating class imbalance. The details are illustrated in Fig. \ref{fig:BBS}. This design ensures that each model update is guided by high-quality, balanced pseudo-labels, thereby improving adaptation robustness in dynamic streams.
    \item We propose Category Prototype-Guided Adaptation (CPGA) to handle continuous domain shifts, aiming to improve pseudo-label quality and stabilize TTA. As shown in Fig. \ref{fig:CPGA}, CPGA dynamically maintains class prototypes during testing and incorporates feature-prototype similarity into model updates, enabling active adaptation in the feature space.
    \item We evaluate our method on three datasets under data streams with coexisting class imbalance and continually changing distributions, demonstrating superior performance over recent state-of-the-art methods.
\end{itemize}

\section{Related Work}
\subsection{Test-Time Adaptation}
Neural networks have become the leading topic of artificial intelligence, and recurrent models \cite{niu2024bidirectional,he2020subtraction,he2023cascade} and ODE-based methods \cite{xu2026cnm,yu2026mobileode,he2024lightweight} have been largely developed for complicate machine learning tasks. However, these models performed not well in dynamic scenarios with continual domain shifts. Unsupervised Domain Adaptation (UDA) aims to train a model using labeled source domain data and generalize it to an unlabeled target domain, addressing the challenge of distribution shift without requiring target domain annotations \cite{udp1, udp2, udp3, udp4}. However, it may be infeasible when training data are unavailable due to privacy or storage concerns. To overcome this limitation, \cite{tent} proposed Test-Time Adaptation (TTA), which focuses on real-time model adjustment during the testing phase using unlabeled target data, without accessing the source data used for training. Early TTA approaches mainly focus on adapting normalization statistics or entropy minimization. For instance, BN-1 \cite{bn1} adapts models by aligning batch normalization statistics to the target domain, which provides a simple yet effective way to partially mitigate distribution shifts. TENT \cite{tent} further updates model parameters by minimizing the prediction entropy on incoming test samples while simultaneously updating batch normalization statistics. In contrast to these partial-update strategies, AdaContrast \cite{adacontrast} introduces a memory module combined with contrastive learning and performs adaptation on the entire network, enabling richer representation learning during test time. Overall, these methods demonstrate that models can effectively adapt to distribution shifts using only unlabeled test data, highlighting the promise of test-time learning as a practical and flexible solution for improving model robustness.

\subsection{Real-World Test-Time Adaptation}
While existing TTA methods have demonstrated strong potential for improving out-of-distribution generalization, most of them are typically developed and evaluated under relatively controlled experimental settings, where the assumptions about the test environment are simplified and the distribution shifts are often limited in scope. Consequently, recent studies on TTA have increasingly focused on extending adaptation methods to more realistic scenarios. Several key challenges have been identified in real-world applications, including adaptation with small-batch or single-sample updates \cite{zhao2024nanoadapt,seto2024realm}, class-imbalanced test data \cite{note}, continuously shifting target domains \cite{cotta,palm,law}, adaptation on noisy data streams \cite{gong2023sotta}, mixtures of multiple distribution shifts \cite{sar}, and constraints imposed by limited computational resources or deployment on edge devices \cite{lame,surgeon}. Empirical evidence shows that these real-world conditions can significantly degrade the performance of existing TTA methods. In this work, we address the combined challenges of continuous domain shifts and severe class imbalance by introducing a dynamic batch-balancing component combined with class prototype guidance.

\subsection{Prototype Learning}
Prototype-based methods represent each class by a compact and representative feature vector, known as a prototype, typically computed as the mean of embedded samples belonging to that class. While the idea of class prototypes has a long history in pattern recognition, Prototypical Networks \cite{snell2017prototypical} popularized its use in modern deep learning, particularly in few-shot learning, by classifying queries based on distances to class prototypes rather than parametric classifiers. Since then, prototype-based methods have been widely adopted in various scenarios, including domain adaptation \cite{jang2023test,sun2024program} and semantic segmentation \cite{sacha2023protoseg,tang2024hunting}, due to their robustness and interpretability. In TTA setting, T3A \cite{T3A} leveraged class prototypes to adjust classifier weights during test time. In contrast, our method utilizes prototype information to guide feature adaptation, rather than confining it to classifier recalibration, and is thus better suited to handling more complex and evolving test-time data distributions.

\begin{figure*}[t]
    \centering
    \includegraphics[width=1\textwidth]{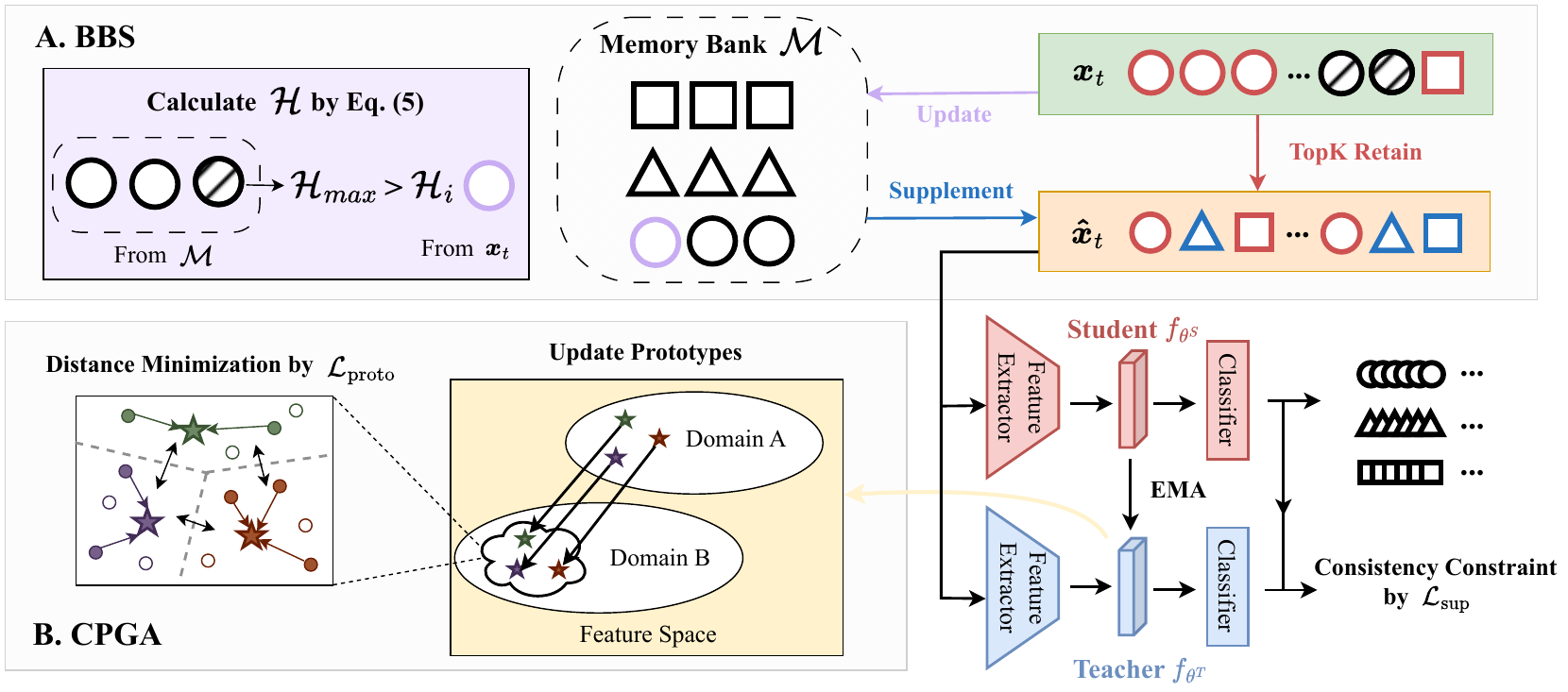}
    \caption{The overall framework of our BP-TTA method. The method addresses class imbalance in streaming data through the introduction of BBS, and mitigates domain shift by aligning cross-domain features using CPGA. During test-time adaptation, we optimize the combination of the consistency loss $\mathcal{L}_{\text{sup}}$ and the prototype loss $\mathcal{L}_{\text{proto}}$.}
    \label{fig:framework}
\end{figure*}

\section{Methods}

\subsection{Problem Definition}

Given a model $f_{\theta}$ pre-trained on a labeled source dataset $\mathcal{D}^{Source} = \{(\boldsymbol{x}^{Source}, \boldsymbol{y}^{Source})\}$, and a sequence of unlabeled target-domain batches provided over time $\mathcal{D}^{Target} = \{\boldsymbol{x}^{Target}\}$. At each time step $t$, the target samples are drawn from a distribution $P^{Target}_{t}(\boldsymbol{x})$, which evolves over time. Moreover, due to temporal dependencies, the target samples are non-independent and identically distributed, and the underlying distribution evolves continuously over time. This setting is motivated by real-world dynamic scenarios, in which distribution shift and temporal correlation among samples often occur simultaneously. 

To formally describe the model architecture, we define $f_{\theta}(\cdot) = h(g(\cdot))$, where $h(\cdot): \mathbb{R}^{B \times D} \rightarrow \mathbb{R}^{B \times C}$ is the linear classifier for the last layer and $g(\cdot): \mathbb{R}^{B \times F \times H \times W} \rightarrow \mathbb{R}^{B \times D}$ is the feature extractor for the rest. Like most mainstream Test-Time Adaptation (TTA) methods, we adopt a Mean Teacher structure \cite{mt}, where a teacher network guides the student network through consistency regularization, effectively smoothing the learning process over time. Both the student model $f_{\theta^S}$ and the teacher model $f_{\theta^T}$ are initialized with the parameters of the pre-trained source model. At each time step $t$, $f_{\theta^S_t}$ receives a batch of target samples $\boldsymbol{x}_t$, produces predictions $f_{\theta^S_t}(\boldsymbol{x}_t)$, and subsequently updates its parameters to adapt to future inputs, i.e., $\theta_t^S \rightarrow \theta^S_{t+1}$. ${\theta^T_t}$ are updated as an Exponential Moving Average (EMA) of the student model parameters. The objective is to improve the performance of the model during inference on the continually evolving target domain in an online manner, without accessing any source data. We propose BP-TTA, which enhances robustness in Practical Test-Time Adaptation (PTTA) through two complementary components. An overview of BP-TTA is illustrated in Fig. \ref{fig:framework}.

\subsection{Batch-Balanced Sampling}
\label{sec:BBS}

In dynamic scenarios with strong temporal dependencies, the class distribution within each incoming batch $\boldsymbol{x}$ is often highly imbalanced. To mitigate class imbalance, we propose a Batch-Balanced Sampling (BBS) component that dynamically reconstructs balanced batches directly from realistic test streams.

Our approach maintains a class-aware memory bank $\mathcal{M}$, which is initially empty and dynamically updated as new batches arrive. Due to the continuously changing test distribution, old samples in $\mathcal{M}$ may lose their relevance or even hinder the model's adaptation, while high-uncertainty samples can generate erroneous gradients that impair learning. To account for these factors, each sample stored in the memory bank $\mathcal{M}$ is associated with a set of attributes $(\hat{y}_i, a_i, \gamma_i, u_i)$
\cite{rotta}. Here, $\hat{y}_i$ is the pseudo-label generated by the teacher model $f_{\theta^T}$, and $a_i$, $\gamma_i$, and $u_i$ denote the age of the sample, the confidence of the pseudo-label, and the uncertainty of the prediction, respectively. Specifically, $\hat{y}_i$, $\gamma_i$, and $u_i$ are computed as follows:
\begin{align}
p(c|x_i) &= \mathrm{Softmax}\big(f_{\theta^T}(x_i)\big), \label{eq:probability} \\
\hat{y}_i &= \arg\max_{c} p(c|x_i), \label{eq:pseudo-label}\\
\gamma_i &= \max_{c} p(c|x_i), \label{eq:confidence} \\
u_i &= - \sum_{c=1}^{C} p(c|x_i) \log(p(c|x_i)), \label{eq:uncertainty}
\end{align}
where $x_i$ denotes the $i$-th test sample in current batch and $C$ is the number of categories. The sample age $a_i$ is initialized to 0 and increases with time $t$. Each incoming sample is selectively inserted into $\mathcal{M}$ according to a selection score that accounts for both temporal recency and uncertainty:
\begin{equation}
\mathcal{H}_i = \exp\Big(\frac{a_i}{K}\Big) + u_i,
\label{eq:score}
\end{equation}
where $K$ is the memory capacity. When the class-specific quota is exceeded, existing samples in the memory bank with higher selection scores are removed to accommodate new samples, ensuring that the memory bank retains more reliable and informative samples for each class.

\begin{algorithm}[t]
\caption{Batch-Balanced Sampling}
\label{alg:balanced_batch_sampling}
\DontPrintSemicolon
\SetAlgoInsideSkip{smallskip}

\textbf{Input:} current batch $\boldsymbol{x} = \{x_i\}_{i=1}^{N}$ and the teacher model $f_{\theta^T}$ \\
\textbf{Define:} number of classes $C$, per-class quota $N_{\text{per-class}}$, memory bank $\mathcal{M} = \{(s^j, \hat{y}^j, a^j, \gamma^j, u^j)\}_{j=1}^{|\mathcal{M}|}$, and its capacity $K$. \\
\textbf{Output:} balanced batch $\boldsymbol{\hat{x}}$\\
\For{$i = 1$ \KwTo $N$}{
    Set $a_i = 0$ and Calculate $p(c|x_i)$, $\hat{y}_i$, $\gamma_i$, $u_i$ by Eq.~\eqref{eq:probability}--\eqref{eq:uncertainty}\;
    $\mathcal{M}_{\hat{y}_i} = \{\, (s^j, \hat{y}^j, a^j, \gamma^j, u^j) \in \mathcal{M}|\hat{y}^j = \hat{y}_i \,\}$\;
    \If{$|\mathcal{M}_{\hat{y}_i}| < N_{\text{per-class}}$}{
        Add $(x_i, \hat{y}_i, a_i, \gamma_i, u_i)$ to $\mathcal{M}$\;
    }
    \Else{
        Calculate $\mathcal{H}_i$ by Eq.~\eqref{eq:score}\;
        Find $(s^j, \hat{y}^j, a^j, \gamma^j, u^j) \in \mathcal{M}_{\hat{y}_i}$ with the highest $\mathcal{H}_s$.\;
        \If{$\mathcal{H}_i < \mathcal{H}_{s}$}{
            Remove $(s^j, \hat{y}^j, a^j, \gamma^j, u^j)$ from $\mathcal{M}$\;
            Add $(x_i, \hat{y}_i, a_i, \gamma_i, u_i)$ to $\mathcal{M}$\;
        }
    }
}
for $j \in \{1, \dots, |\mathcal{M}|\}$, $a^j \leftarrow a^j + 1$\;
$\boldsymbol{\hat{x}} \leftarrow \emptyset$\;
\For{$c = 1$ \KwTo $C$}{
    $\mathcal{B}_c = \{ (x_i, \hat{y}_i, \gamma_i)\mid \hat{y}_i = c, x_i \in \boldsymbol{x} \}$\;

    \If{$|\mathcal{B}_c| > N_{\text{per-class}}$}{
        $\mathcal{B} \leftarrow \text{TopK}(\mathcal{B}_c, N_{\text{per-class}}, \text{key}=\gamma)$\;
    }
    \Else{
        $\mathcal{B}_{\mathcal{M}} = \{ (x_i, \hat{y}_i, \gamma_i)\mid \hat{y}_i = c, x_i \in \mathcal{M} \}$\;
        $\mathcal{B} \leftarrow \mathcal{B}_c \cup \text{TopK}(\mathcal{B}_{\mathcal{M}}, N_{\text{per-class}} - |\mathcal{B}_c|, \text{key}=\gamma)$\;

        \While{$|\mathcal{B}| < N_{\text{per-class}}$}{
            Add samples from $\mathcal{B}$ with highest $\gamma$\;
        }
    }
    $\hat{\boldsymbol{x}} \leftarrow \hat{\boldsymbol{x}} \cup
    \{ x_i \mid (x_i, \hat{y}_i, \gamma_i) \in \mathcal{B} \}$\;
    \If{$|\hat{\boldsymbol{x}}| \ge N$}{
        \textbf{break}\;
    }
}

\Return $\boldsymbol{\hat{x}}$\;
\end{algorithm}

Using an algorithm described in Algorithm~\ref{alg:balanced_batch_sampling}, we dynamically reconstruct a balanced batch $\hat{\boldsymbol{x}}_t$ with exactly $N_{\text{per-class}}$ samples per class for model adaptation at each time step. We set $N_{\text{per-class}} = \frac{N}{C}$ by default. When the number of categories is extremely large, we instead use $N_{\text{per-class}} = \frac{N}{U_t}$, where $U_t$ is the number of predicted pseudo-labels at time step $t$. By eliminating redundant majority-class samples and adding high-quality memory samples to underrepresented minority classes, our method successfully reduces class imbalance. Specifically, BBS first partitions the current batch $\boldsymbol{x}_t$ by pseudo-labels, grouping samples into class-specific sets. It then adjusts the processing approach for each class based on the representation of the class in the incoming batch. We choose the top-$N_{\text{per-class}}$ highest-confidence samples for majority classes that surpass the target count, thereby downsampling without sacrificing quality. We retrieve additional samples from the corresponding class subset in the memory bank, giving priority to those with high confidence scores, for minority classes that fall short of the target. We take a direct sample from the memory bank if a class is not present in the current batch. To guarantee that each selected class reaches $N_{\text{per-class}}$ samples in the final balanced batch, we replicate the highest-confidence sample in rare cases where even memory augmentation is insufficient.

\textbf{Innovations of BBS}.
RoTTA \cite{rotta} proposed Category-balanced Sampling with Timeliness and Uncertainty (CSTU), which also leverages a class-balanced memory bank to mitigates test distribution shift in PTTA for effective model adaptation. Compared to CSTU, BBS brings clear and substantial improvements. First, BBS adopts a simpler score formulation that jointly considers temporal recency and prediction uncertainty, as defined in Eq.~\eqref{eq:score}, while remaining equally effective in practice. Second, although CSTU takes both timeliness and category balance into account, its batches are composed solely of samples retrieved from the memory bank. In scenarios with frequent domain changes, CSTU still introduces temporal delay, preventing the constructed batches from faithfully reflecting the current target-domain distribution. In contrast, BBS directly operates on the incoming test batch, selectively filtering redundant majority-class samples and supplementing underrepresented minority classes with high-quality historical samples from the memory bank. The resulting balanced batch not only provides equal learning opportunities for all classes but also better reflects the distribution of the current test stream, leading to more stable and robust adaptation.

\subsection{Category Prototype-Guided Adaptation}
\label{sec:CPGA}

Under continuous domain shifts, pseudo-labels generated solely by entropy minimization or model predictions tend to become increasingly noisy and unreliable. We propose Category Prototype-Guided Adaption (CPGA) to stabilize TTA under persistent domain shifts. It dynamically maintains class prototypes using an exponential moving average and leverages sample-prototype similarity to guide proactive adaptation of the feature space. This design promotes tighter intra-class clustering and more discriminative inter-class separation, thereby improving pseudo-label quality under persistent domain shifts.

To preserve stable class-level structures under continuous domain shifts, we maintain a set of class prototypes $\{\boldsymbol{p}_c \in \mathbb{R}^D\}_{c=1}^{C}$, each representing the centroid of a class in the feature space and providing class-level guidance. Leveraging the discriminative class knowledge already encoded in the source-trained model's classifier, we initialize these prototypes as follows:
\begin{equation}
\boldsymbol{p}_c^{(0)} = \mathbf{W}_c, \quad 1 \leq c \leq C,
\end{equation}
where $\mathbf{W}_c \in \mathbb{R}^D$ is the $c$-th row of the classifier weight matrix and $D$ is the dimension of feature embeddings. During TTA, these prototypes must track the evolving feature distribution to remain effective under continuous domain shift. For each balanced batch $\hat{\boldsymbol{x}}_t = \{\hat{x}_i\}_{i=1}^{N} $ constructed in Section~\ref{sec:BBS}, we extract features $\boldsymbol{z}_i = g_t(\hat{x}_i)$ and calculate the pseudo-label $\hat{y}_i$ by Eq.~\eqref{eq:pseudo-label} for $1 \leq i \leq N$. To maintain prototype quality and prevent error accumulation from noisy pseudo-labels, we selectively update prototypes using only high-confidence samples through a momentum-based mechanism:
\begin{equation}
\boldsymbol{p}_c^{(t+1)} = \alpha \boldsymbol{p}_c^{(t)} + (1 - \alpha) \boldsymbol{z}_i, \quad \text{if } \hat{y}_i = c \text{ and } \gamma_i \geq \tau,
\label{eq:proto_update}
\end{equation}
where $\alpha\in[0,1]$ is a smoothing factor, $\hat{y}_i$ is the teacher's pseudo-label, $\gamma_i$ is the confidence score calculated by Eq.~\eqref{eq:confidence}, and $\tau$ is a confidence threshold. A lower momentum value enables prototypes to respond promptly to distribution changes while maintaining stability against noisy updates. With the updated prototypes, we enforce class-level consistency in the feature space through a prototype-based supervision loss. The key idea is to measure how well each sample's feature aligns with different class prototypes and use this alignment to guide the learning process. Given a sample $x_i$ with extracted feature $\boldsymbol{z}_i \in \mathbb{R}^{D}$, we compute cosine similarity between it and its assigned class prototype:
\begin{equation}
\text{sim}(\boldsymbol{z}_i, \boldsymbol{p}_{\hat{y}_i}) = \frac{\boldsymbol{z}_i \cdot \boldsymbol{p}_{\hat{y}_i}}{\|\boldsymbol{z}_i\| \|\boldsymbol{p}_{\hat{y}_i}\|},
\end{equation}
where $\text{sim}(\boldsymbol{z}_i, \boldsymbol{p}_{\hat{y}_i}) \in [-1, 1]$ measures similarity between sample $\hat{x}_i$ and its assigned class prototype. Higher similarity indicates stronger alignment between the feature and its target class representation. 
To ensure that prototype updates are guided by reliable supervision, we compute the prototype loss only on high-confidence samples, defined as $\mathcal{S} = \{i|\gamma_i \geq \tau\}$. The loss is expressed as minimizing the distance between features and their corresponding prototypes:
\begin{equation}
\mathcal{L}_{\text{proto}} = \frac{1}{|\mathcal{S}|} \sum_{i \in \mathcal{S}} \left(1 - \text{sim}(\boldsymbol{z}_i, \boldsymbol{p}_{\hat{y}_i})\right).
\end{equation}

\textbf{Innovations of CPGA}.
Prototype-based methods update class prototypes during testing and leverage them to facilitate classification. Early approaches typically do not perform explicit alignment in the feature space of the feature extractor. For example, T3A \cite{T3A} updates class prototypes based on the support set of each category and subsequently adjusts classifier weights according to these prototypes during the test time. TAST \cite{jang2023test} further refines the support set by incorporating nearest-neighbor information, thereby mitigating the impact of noisy pseudo-labels. More recently, TSD \cite{TSD} introduces Memorized Spatial Local Clustering to optimize alignment in the feature space of the feature extractor. However, it still relies on local neighborhood constraints and incurs considerable storage overhead for sample features. We propose the CPGA, which dynamically maintains class prototypes to adapt continual domain shift and employs a prototype alignment loss to directly guide features toward their corresponding class prototypes. This encourages a more discriminative feature space characterized by increased intra-class compactness and enhanced inter-class separability.

\subsection{The Overall Optimization Objective}
The overall optimization objective consists of two parts that jointly guide the adaptation process. For the balanced batch constructed at each time step, we first apply strong data augmentations and enforce a prediction consistency constraint between the student and teacher models:
\begin{equation}
\mathcal{L}_{\text{sup}} = \frac{1}{N} \sum_{i=1}^{N} H\big(p_{\theta^{\text{S}}}(y \mid \text{Aug}(x_i)),\, p_{\theta^{\text{T}}}(y \mid x_i)\big),
\end{equation}
where $H(\cdot, \cdot)$ denotes a consistency loss, and $\text{Aug}(\cdot)$ applies data augmentation. This loss stabilizes the student model by requiring its predictions under perturbations to match the teacher’s outputs, thereby improving robustness under distribution shift.

To further enhance class-level consistency, we incorporate the prototype loss introduced in Section~\ref{sec:CPGA}, which encourages sample features to align with their corresponding class prototypes. The total loss is defined as:
\begin{equation}
\mathcal{L}_{\text{total}} = \mathcal{L}_{\text{sup}} + \lambda \mathcal{L}_{\text{proto}},
\label{eq:loss}
\end{equation}
where $\lambda $ is a regularisation coefficient.

Student model parameters are updated via gradient descent with gradient clipping to prevent instability:
\begin{equation}
\theta_{t+1}^S \leftarrow \theta_t^S - \eta \nabla_{\theta_t^S} \mathcal{L}_{\text{total}}, \quad \|\nabla_{\theta_t^S} \mathcal{L}_{\text{total}}\| \leq \epsilon,
\end{equation}
where $\eta$ is the learning rate and $\epsilon$ is the clipping threshold. The teacher model is updated via exponential moving average:
\begin{equation}
\theta^T_{t+1} \leftarrow (1 - \nu) \theta^T_t + \nu \theta_{t+1}^S,
\end{equation}
where $\nu$ is the EMA momentum.

\section{Experiments}

\subsection{Experiment Settings}
\subsubsection{Datasets}
We follow the standard evaluation benchmarks introduced in \cite{cotta} and evaluate our proposed method on CIFAR10-C, CIFAR100-C, and ImageNet-C. These datasets are built upon the corruption protocols proposed in \cite{hendrycks2019benchmarking}. Each dataset contains an image set of 15 corruption style including gaussian noise, shot noise, impulse noise, defocus blur, glass blur, motion blur, zoom blur, snow, frost, fog, brightness, contrast, elastic, pixelated, and jpeg. We adopt the same test-time augmentation strategy as CoTTA \cite{cotta}, applying transformations such as color jitter, gaussian blur, gaussian noise, random flips, and random affine operations to the input.

\begin{figure}[t]
    \centering
    \resizebox{\columnwidth}{!}{
    \includegraphics{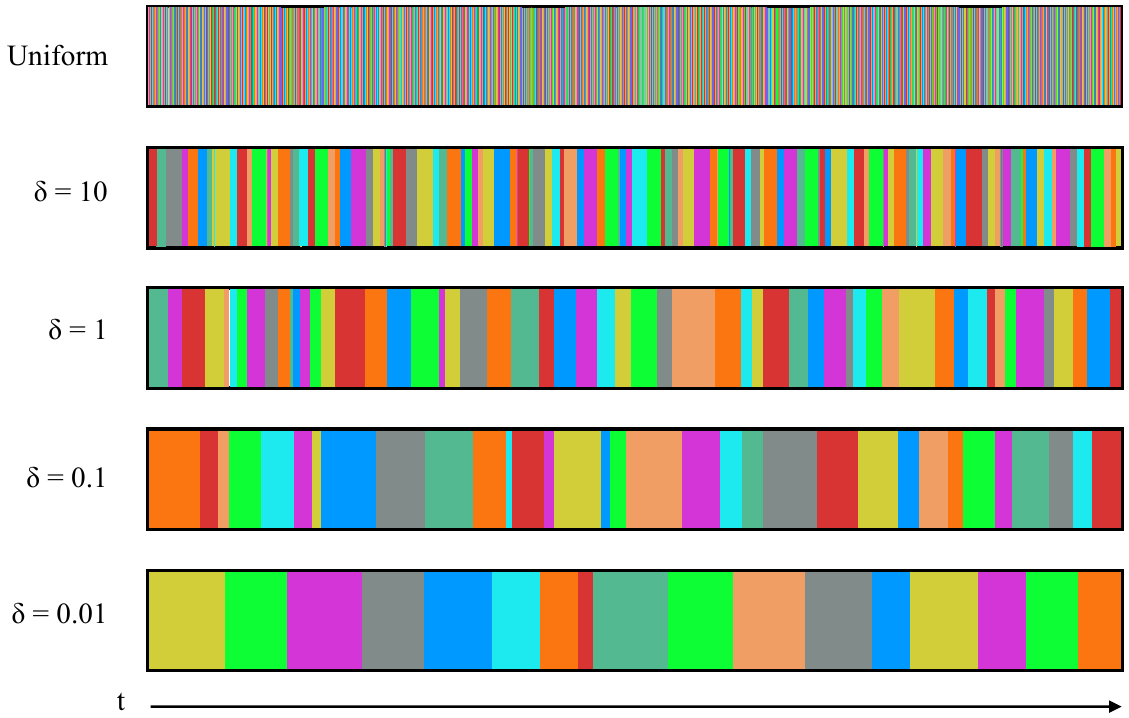}
    }
    \caption{Illustration of class-imbalanced test streams generated by different Dirichlet parameters $\delta$ on CIFAR10-C. Each color represents a different category.}
    \label{fig:dirichlet}
\end{figure}

\subsubsection{Implementation and setup} Following common practice in Test-Time Adaptation (TTA), we adopt WideResNet-28 \cite{zagoruyko2016wide} for CIFAR10-C and ResNeXt-29 \cite{xie2017aggregated} for CIFAR100-C, both sourced from the RobustBench benchmark \cite{croce2robustbench}. For ImageNet-C, we use a ResNet-50 \cite{he2016deep} model provided by the official torchvision library. All evaluations are conducted under the highest corruption severity (level 5). During the TTA process, the memory bank capacity $K$ is set to 100, 200, and 1000 for CIFAR10-C, CIFAR100-C, and ImageNet-C datasets, respectively. EMA momentum $\nu$ is fixed to 0.001 for all experiments. Smoothing factor $\alpha$ is set to 0.3 for CIFAR10-C and 0.5 for CIFAR100-C and ImageNet-C. Threshold $\tau$ is set to 0.8 for CIFAR10-C and CIFAR100-C, and 0.5 for ImageNet-C. $\lambda $ and $\epsilon $ are set to 10 and 1 throughout. To construct the PTTA test data stream, we sample from the Dirichlet distribution with concentration parameter $\delta$ configured as 0.1, 0.01, and 0.001 respectively. Fig. \ref{fig:dirichlet} illustrates the class distribution of test streams generated with different $\delta$ on CIFAR10-C. A smaller $\delta$ leads to a more imbalanced class distribution, while a larger $\delta$ results in a more balanced distribution.

For optimization, we adopt Adam optimizer with a learning rate $\eta $ of 1e-3 and $\beta=0.9$, and the batch size is consistently set to 64 across all datasets and methods. All experiments were implemented with PyTorch in Python 3.9 on GeForce 4090 GPUs.

\subsubsection{Baselines} We evaluate our proposed BP-TTA against several representative TTA methods, including TENT \cite{tent}, CoTTA \cite{cotta}, EATA \cite{eata}, SAR \cite{sar}, PALM \cite{palm}, and SURGEON \cite{surgeon}. In addition, we include comparisons with TTA methods specifically designed for class-imbalanced scenarios, such as LAME \cite{lame}, RoTTA \cite{rotta}, and DA-TTA \cite{da-tta}. For a fair evaluation, all baselines are re-implemented with parameters as mentioned in their paper and tested under identical experimental settings. In all result tables, the best scores are
highlighted in bold, and the second-best scores are underlined.

\begin{table*}[!t]
  \centering
  \caption[b]{Average classification error (\%) on CIFAR10-C, CIFAR100-C, and ImageNet-C under the PTTA setting}
  \resizebox{\textwidth}{!}{
    \begin{tabular}{c|cc|ccccccccccccccc|c}
    \toprule
    & \multicolumn{2}{c|}{Time} & \multicolumn{15}{c|}{$t \xrightarrow{\hspace{16cm}}$} \\
    \midrule
    & \multicolumn{2}{c|}{Method} & Brigh. & Zoom & Contr. & Moti. & Snow & Fog & Defoc. & Jpeg & Glass & Frost & Pixel. & Gauss. & Elast. & Shot & Impul. & Avg. \\
    \midrule
    \multirow{11}{*}{\rotatebox{90}{CIFAR10-C}} 
    & \multicolumn{2}{c|}{Source \cite{zagoruyko2016wide}} & 9.31 & 42.02 & 46.70 & 34.75 & 25.07 & 26.01 & 46.94 & 30.30 & 54.32 & 41.30 & 58.45 & 72.33 & 26.59 & 65.73 & 72.92 & 43.52 \\
    & \multicolumn{2}{c|}{TENT \cite{tent}} & 71.85 & 74.22 & 77.95 & 80.49 & 83.81 & 84.75 & 86.16 & 87.91 & 87.66 & 87.25 & 87.96 & 88.38 & 88.28 & 89.21 & 89.44 & 84.35 \\
    & \multicolumn{2}{c|}{CoTTA \cite{cotta}} & 73.69 & 76.88 & 81.03 & 79.96 & 80.92 & 80.71 & 81.37 & 81.46 & 82.94 & 82.16 & 82.87 & 82.76 & 83.47 & 82.72 & 84.11 & 81.14 \\
    & \multicolumn{2}{c|}{EATA \cite{eata}} & 72.08 & 72.71 & 72.19 & 73.26 & 74.47 & 72.90 & 73.30 & 77.47 & 80.24 & 72.55 & 77.29 & 77.15 & 77.13 & 76.45 & 78.98 & 75.21 \\
    & \multicolumn{2}{c|}{SAR \cite{sar}} & 70.72 & 73.92 & 73.18 & 73.50 & 75.33 & 72.71 & 73.46 & 78.61 & 80.16 & 74.84 & 76.29 & 77.42 & 79.00 & 77.86 & 80.44 & 75.83 \\
    & \multicolumn{2}{c|}{PALM \cite{palm}} & 71.67 & 73.61 & 72.64 & 74.38 & 75.77 & 74.40 & 75.15 & 79.50 & 82.21 & 77.73 & 79.81 & 81.30 & 82.01 & 82.52 & 85.07 & 77.85 \\
    & \multicolumn{2}{c|}{SURGEON \cite{surgeon}} & 77.47 & 76.82 & 79.60 & 76.61 & 81.69 & 79.18 & 79.93 & 81.96 & 83.36 & 84.52 & 83.28 & 86.20 & 86.80 & 87.21 & 88.03 & 82.18 \\
    & \multicolumn{2}{c|}{LAME \cite{lame}} & 3.77 & 7.97 & 41.00 & 5.47 & 6.07 & 7.22 & 24.68 & 5.04 & 41.91 & 17.79 & 50.59 & 78.32 & 5.99 & 69.92 & 62.28 & 28.53 \\
    & \multicolumn{2}{c|}{RoTTA \cite{rotta}} & 10.00 & 17.98 & 20.70 & 20.36 & 23.76 & 17.61 & 15.84 & 30.49 & 40.71 & 18.80 & 30.80 & 32.13 & 32.25 & 26.73 & 35.06 & \underline{24.88} \\
    & \multicolumn{2}{c|}{DA-TTA \cite{da-tta}} & 20.53 & 17.72 & 12.37 & 20.44 & 25.75 & 16.60 & 20.38 & 33.10 & 40.26 & 18.89 & 31.93 & 32.10 & 31.15 & 34.43 & 43.24 & 26.59 \\
    & \multicolumn{2}{c|}{BP-TTA (ours)} & 8.68 & 18.57 & 13.16 & 16.39 & 18.67 & 16.55 & 15.17 & 27.49 & 36.12 & 15.87 & 26.95 & 29.21 & 26.40 & 25.18 & 32.55 & \textbf{21.80} \\
    \midrule
    \multirow{11}{*}{\rotatebox{90}{CIFAR100-C}}
    & \multicolumn{2}{c|}{Source \cite{xie2017aggregated}} & 29.52 & 28.79 & 55.09 & 30.81 & 39.47 & 50.30 & 29.36 & 41.22 & 54.09 & 45.82 & 74.71 & 72.98 & 37.21 & 68.00 & 39.36 & 46.45 \\
    & \multicolumn{2}{c|}{TENT \cite{tent}} & 82.99 & 96.10 & 98.05 & 97.44 & 97.61 & 97.55 & 97.94 & 98.05 & 97.99 & 98.20 & 98.40 & 98.23 & 98.45 & 98.82 & 99.02 & 96.99 \\
    & \multicolumn{2}{c|}{CoTTA \cite{cotta}} & 77.72 & 76.35 & 79.66 & 80.15 & 81.26 & 82.49 & 80.65 & 81.67 & 82.95 & 82.08 & 80.67 & 82.78 & 83.78 & 83.53 & 83.60 & 81.29 \\
    & \multicolumn{2}{c|}{EATA \cite{eata}} & 77.54 & 75.89 & 77.83 & 78.20 & 80.11 & 81.84 & 77.23 & 81.69 & 82.13 & 79.10 & 79.01 & 82.82 & 80.15 & 80.41 & 81.29 & 79.68 \\
    & \multicolumn{2}{c|}{SAR \cite{sar}} & 79.03 & 84.75 & 90.02 & 93.33 & 96.44 & 97.07 & 97.18 & 94.46 & 85.19 & 88.29 & 92.73 & 95.79 & 96.31 & 97.52 & 97.78 & 92.39 \\
    & \multicolumn{2}{c|}{PALM \cite{palm}} & 77.10 & 78.31 & 82.15 & 83.63 & 87.70 & 91.37 & 93.48 & 95.38 & 96.98 & 97.78 & 98.24 & 98.34 & 98.37 & 98.47 & 98.68 & 91.73 \\
    & \multicolumn{2}{c|}{SURGEON \cite{surgeon}} & 82.02 & 81.07 & 81.84 & 80.77 & 85.33 & 85.31 & 87.04 & 89.38 & 90.94 & 94.49 & 93.51 & 95.99 & 96.03 & 96.27 & 96.53 & 89.10 \\
    & \multicolumn{2}{c|}{LAME \cite{lame}} & 40.08 & 65.16 & 89.08 & 74.32 & 63.02 & 92.60 & 70.60 & 53.59 & 89.11 & 84.64 & 92.21 & 89.63 & 76.39 & 89.27 & 67.32 & 75.80 \\
    & \multicolumn{2}{c|}{RoTTA \cite{rotta}} & 31.64 & 34.94 & 42.79 & 34.63 & 37.22 & 41.68 & 29.51 & 41.25 & 42.66 & 35.81 & 37.62 & 41.75 & 38.89 & 40.68 & 46.17 & \underline{38.48} \\
    & \multicolumn{2}{c|}{DA-TTA \cite{da-tta}} & 26.63 & 28.20 & 77.81 & 94.54 & 35.19 & 40.49 & 28.76 & 58.30 & 41.83 & 34.23 & 32.41 & 41.93 & 37.88 & 38.64 & 38.58 & 43.69 \\
    & \multicolumn{2}{c|}{BP-TTA (ours)} & 25.71 & 30.49 & 31.44 & 31.18 & 34.35 & 40.28 & 29.48 & 40.83 & 42.13 & 34.00 & 39.03 & 44.81 & 39.31 & 41.40 & 45.10 & \textbf{36.64} \\
    \midrule
    \multirow{11}{*}{\rotatebox{90}{ImageNet-C}}
    & \multicolumn{2}{c|}{Source \cite{he2016deep}} & 41.11 & 78.03 & 94.62 & 85.45 & 83.49 & 76.38 & 81.76 & 67.42 & 89.63 & 77.43 & 79.11 & 97.11 & 82.97 & 96.45 & 97.55 & 81.90 \\
    & \multicolumn{2}{c|}{TENT \cite{tent}} & 45.81 & 67.20 & 92.60 & 96.83 & 98.86 & 99.26 & 99.54 & 99.52 & 99.58 & 99.62 & 99.56 & 99.66 & 99.68 & 99.66 & 99.68 & 93.14 \\
    & \multicolumn{2}{c|}{CoTTA \cite{cotta}} & 98.92 & 98.80 & 99.72 & 99.52 & 99.46 & 99.54 & 99.70 & 99.30 & 99.72 & 99.60 & 99.36 & 99.38 & 99.30 & 99.32 & 99.46 & 99.40 \\
    & \multicolumn{2}{c|}{EATA \cite{eata}} & 45.30 & 70.06 & 87.66 & 78.57 & 72.81 & 61.22 & 88.02 & 66.99 & 87.66 & 73.82 & 60.57 & 86.86 & 64.19 & 86.16 & 86.36 & 74.42 \\
    & \multicolumn{2}{c|}{SAR \cite{sar}} & 44.54 & 64.77 & 81.48 & 72.69 & 71.93 & 61.38 & 80.48 & 64.47 & 81.50 & 75.10 & 63.70 & 79.47 & 67.13 & 77.47 & 78.45 & 70.97 \\
    & \multicolumn{2}{c|}{PALM \cite{palm}} & 44.44 & 67.05 & 84.13 & 74.26 & 67.84 & 57.12 & 82.34 & 62.32 & 81.94 & 69.70 & 58.55 & 80.32 & 60.85 & 78.35 & 78.05 & 69.82 \\
    & \multicolumn{2}{c|}{SURGEON \cite{surgeon}} & 84.99 & 81.84 & 80.08 & 83.47 & 81.44 & 74.68 & 66.67 & 68.36 & 69.38 & 60.79 & 48.03 & 77.77 & 60.63 & 59.01 & 61.40 & 70.57 \\
    & \multicolumn{2}{c|}{LAME \cite{lame}} & 52.41 & 97.43 & 99.40 & 99.52 & 99.92 & 99.32 & 90.95 & 91.87 & 99.12 & 98.31 & 97.31 & 99.82 & 99.78 & 99.74 & 99.80 & 94.98 \\
    & \multicolumn{2}{c|}{RoTTA \cite{rotta}} & 35.93 & 64.37 & 85.85 & 75.54 & 66.55 & 55.64 & 80.40 & 58.03 & 84.05 & 68.74 & 52.97 & 86.18 & 61.16 & 78.71 & 75.50 & 68.64 \\
    & \multicolumn{2}{c|}{DA-TTA \cite{da-tta}} & 43.96 & 77.49 & 78.49 & 72.89 & 67.52 & 51.12 & 78.83 & 57.80 & 80.38 & 68.28 & 51.93 & 80.80 & 56.90 & 80.16 & 78.81 & \underline{68.36} \\
    & \multicolumn{2}{c|}{BP-TTA (ours)} & 37.68 & 64.00 & 82.76 & 75.02 & 64.67 & 52.53 & 80.32 & 58.21 & 83.27 & 63.98 & 52.15 & 78.33 & 60.51 & 74.92 & 73.23 & \textbf{66.77} \\
    \bottomrule
    \end{tabular}
  }
  \label{tab:sota}
\end{table*}

\subsection{Results}
Table~\ref{tab:sota} presents the experimental results of BP-TTA and other TTA baseline methods for adaptation on sample streams with domain shift and class imbalance. The results show that our proposed BP-TTA method achieves the best performance across all baselines on the three datasets, outperforming the second-best method by margins of 3.08\%, 1.84\%, and 1.59\%, respectively. It is observed that many previous methods exhibit poorer results under the Practical Test-Time Adaptation (PTTA) setting, for they tend to overfit local distributions when confronted with class-imbalanced test streams. Notably, although LAME shows favorable performance advantages on CIFAR10-C and even achieves optimal results on some target domains without updating the model's parameters, it suffers from substantial performance degradation on CIFAR100-C and ImageNet-C. On the contrary, BP-TTA maintains stable adaptation capabilities across all target domains and achieves the best overall average performance.

\begin{table}[t]
    \centering
    \caption{Average classification error (\%) on DomainNet-126 under the PTTA setting with the domain shift sequence $ \text{Real} \rightarrow \text{Painting} \rightarrow \text{Clipart} $}
    \label{tab:domainnet126}
    \begin{tabular}{l|ccc|c}
    \toprule[1.2pt] 
    Method & Real & Painting & Clipart & Avg. \\
    \midrule
    Source    & 40.80  & 50.20  & 45.38 & 43.91 \\
    TENT \cite{tent}     & 98.17 & 99.15 & 99.01 & 98.55 \\
    CoTTA \cite{cotta}     & 98.35  & 98.62  & 98.43 & 98.43 \\
    EATA \cite{eata}     & 82.60  & 77.02  & 72.17 & 79.54 \\
    SAR \cite{sar}    & 87.55  & 90.51  & 88.36 & 88.43 \\
    PALM \cite{palm}     & 96.26  & 98.91  & 99.03 & 97.37 \\
    LAME \cite{lame}      & 32.34  & 56.36  & 49.34 & 41.11 \\
    RoTTA \cite{rotta}    & 33.86  & 42.99  & 44.60 & \underline{37.87}\\
    DA-TTA \cite{da-tta}    & 34.54  & 75.81  & 97.70 & 54.93 \\
    BP-TTA (ours) & 33.06  & 41.81  & 45.35 & \textbf{37.21}\\
    \bottomrule[1.2pt] 
    \end{tabular}
\end{table}

\subsection{Additional experiment on DomainNet-126}
To validate the effectiveness of our approach under domain shifts not induced by corruption, we further conduct experiments on the DomainNet-126 dataset~\cite{domainnet126}. DomainNet-126 comprises 126 classes across four domains (real, clipart, painting, and sketch), each containing 18,523 images. We continually adapt a ResNet-50 model pre-trained on the sketch domain to correlatively sampled test streams of the rest domains. The architecture and pre-trained weights are consistent with those used in~\cite{adacontrast}. Notably, this architecture employs weight normalization, which decouples the weights into magnitude and direction components. This reparameterization modifies the gradient flow and leads to incompatibility with the sparsity mechanism and backpropagation process of SURGEON~\cite{surgeon}. Therefore, we exclude SURGEON from this experiment. As shown in Table~\ref{tab:domainnet126}, BP-TTA achieves the best performance, demonstrating its strong capability in handling more challenging domain shifts.

\subsection{Ablation Studies and Analysis}

\subsubsection{Effect of each component}
We conduct an ablative study to further investigate the effectiveness of individual components. As shown in Table~\ref{table:ablation}, using BBS alone significantly reduces error rates over the source model, demonstrating the effectiveness of balanced sampling in mitigating class imbalance during adaptation. CPGA alone underperforms the source model on CIFAR10-C and CIFAR100-C, since severe class imbalance in the test stream causes biased prototype updates, leading to error accumulation. This is mainly because the highly imbalanced test stream leads to biased and noisy category prototypes. The joint use of BBS and CPGA outperforms either component alone, as BBS provides balanced batches for reliable prototype updates, while CPGA further exploits these stable prototypes for effective adaptation.

\subsubsection{Effect of batch size}
\begin{figure}[t]
    \centering
    \subfloat[\label{fig:batchsize}]{%
        \includegraphics[width=0.48\columnwidth]{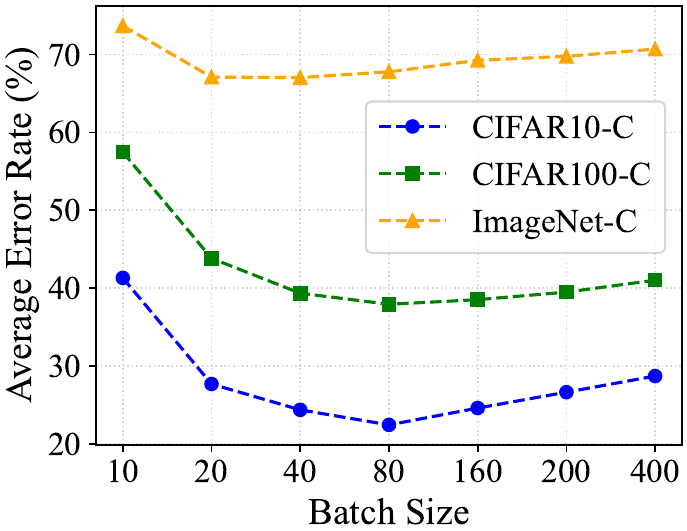}%
    }\hfill
    \subfloat[\label{fig:tau_alpha}]{%
        \includegraphics[width=0.48\columnwidth]{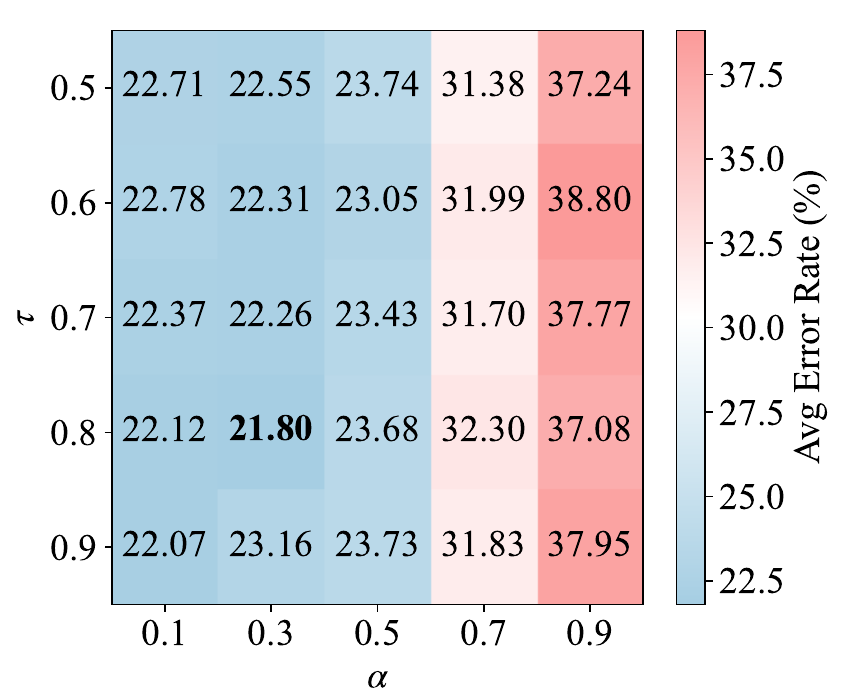}%
    }
    \caption{Effect of batch size on the performance of BP-TTA across three datasets and the effect of hyperparameters $\tau$ and $\alpha$ on CIFAR10-C.}
\end{figure}

As shown in Fig. \ref{fig:batchsize}, batch size has a noticeable impact on the performance of BP-TTA. Compared to very small batch sizes, increasing the batch size significantly reduces the classification error on three datasets. However, when the batch size becomes excessively large, the performance tends to degrade. This is because a small batch size introduces large noise when updating the student network and category prototypes, making the adaptation unstable, while an overly large batch size slows down the model's response to distribution shifts and may generate inaccurate pseudo-labels from an outdated teacher network. Therefore, choosing a moderate batch size can balance stability and response speed, achieving the best performance.

\begin{table}[t]
    \centering
    \caption{Ablation study of average classification error (\%) under the PTTA setting}
    \setlength{\tabcolsep}{1.5mm}{
    \begin{tabular}{l|ccc}
        \toprule[1.2pt]
        Method & CIFAR10-C & CIFAR100-C & ImageNet-C  \\
        \midrule
         Source & 43.52 & 46.45 & 81.90 \\
         BBS & 25.57 & 37.84 & 67.78  \\
         CPGA & 50.92 & 66.04 & 68.19 \\
         BBS+CPGA & \bf 21.80 & \bf 36.64 & \bf 66.77 \\
    \bottomrule[1.2pt]
    \end{tabular}
    }
    \label{table:ablation}
\end{table}

\subsubsection{Effect of hyperparameters $\tau$ and $\alpha$}
We analyze the impact of the confidence threshold $\tau$ and smoothing factor $\alpha$ in Eq.~\eqref{eq:proto_update} on the performance of BP-TTA. Taking CIFAR10-C as an example, Fig. \ref{fig:tau_alpha} presents the average error rates under different combinations of $\tau$ and $\alpha$. We observe that when the value of $\tau$ ranges from 0.5 to 0.9, the results do not vary too much, indicating that the model does not heavily rely on precise parameter tuning. Regarding the smoothing factor $\alpha$, lower values generally yield better results, as they allow class prototypes to respond more quickly to continual domain shifts, while large values may slow down the adaptation process and reduce the effectiveness of our method.

\begin{table}[t]
  \centering
  \caption{Average classification error (\%) on CIFAR10-C, CIFAR100-C, and ImageNet-C for different values of $\lambda$}
  \begin{tabular}{c|ccccc}
      \toprule
      \textbf{$\lambda$} & \textbf{0} & \textbf{0.1} & \textbf{1} & \textbf{10} & \textbf{100} \\
      \midrule
      CIFAR10-C   & 25.57 & 23.37 & 22.53 & \textbf{21.80} & 22.66 \\
      CIFAR100-C  & 37.84 & 36.80 & 36.83 & \textbf{36.64} & 37.65 \\
      ImageNet-C  & 67.78 & 66.91 & 66.84 & \textbf{66.77} & 67.05 \\
      \bottomrule
  \end{tabular}
  \label{tab:lambda}
\end{table}

\subsubsection{Sensitivity to weighting factor}
We conduct a detailed hyperparameter search for weighting factor $\lambda$, i.e., $\lambda \in \{0, 0.1, 1, 10, 100\}$, to investigate the sensitivity of $\lambda$ in Eq.~\eqref{eq:loss}. $\lambda$ is the hyper-parameter balancing $\mathcal{L}_{\text{sup}}$ and $\mathcal{L}_{\text{proto}}$. $\mathcal{L}_{\text{sup}}$ provides direct supervision for samples with high-confidence pseudo-labels, and $\mathcal{L}_{\text{proto}}$ encourages feature representations to align with their corresponding class prototypes, therefore promoting intra-class clustering and inter-class separation. Table~\ref{tab:lambda} reports the average classification error rates obtained with different values of $\lambda$. From the results, we observe that setting $\lambda$ to 10 consistently yields the best performance across all three datasets, demonstrating the effectiveness of the prototype-guided loss.

\subsubsection{Effect of domain change frequency}
To further investigate how different methods respond to varying frequencies of domain changes, we introduce a parameter $n$ to explicitly control the frequency of domain shifts in the test data stream. Specifically, $n$ denotes the number of segments into which the sample sequence of each domain is divided, such that increasing $n$ results in more frequent domain transitions throughout the test stream. When $n=1$, the setting degenerates to the original domain shift scenario where each domain appears only once in sequence. Table~\ref{tab:frequency} presents the average classification errors on ImageNet-C. As can be seen, RoTTA suffers a substantial performance drop when $n$ increases, indicating its sensitivity to frequent domain changes. Meanwhile, BP-TTA exhibits a slower performance degradation, which proves BBS can effectively alleviate the negative impact of frequent domain shifts.

\begin{table}[t]
  \centering
  \caption{Average classification error (\%) on ImageNet-C for different values of $n$}
  \begin{tabular}{c|ccccc}
      \toprule
      \textbf{$n$} & \textbf{1} & \textbf{2} & \textbf{4} & \textbf{6} & \textbf{8}  \\
      \midrule
      LAME \cite{lame} & 94.98 & 94.85 & 95.12 & 95.33 & 95.32 \\
      RoTTA \cite{rotta}  & \underline{68.64} & \underline{71.30} & \underline{75.14} & \underline{76.97} & \underline{78.61} \\
      BP-TTA (ours)  & \textbf{66.77} & \textbf{69.30} & \textbf{72.24} & \textbf{73.16} & \textbf{74.88} \\
      \bottomrule
  \end{tabular}
  \label{tab:frequency}
\end{table}

\subsubsection{t-SNE visualization analysis}
\begin{figure}[t]
    \includegraphics[width=\columnwidth]{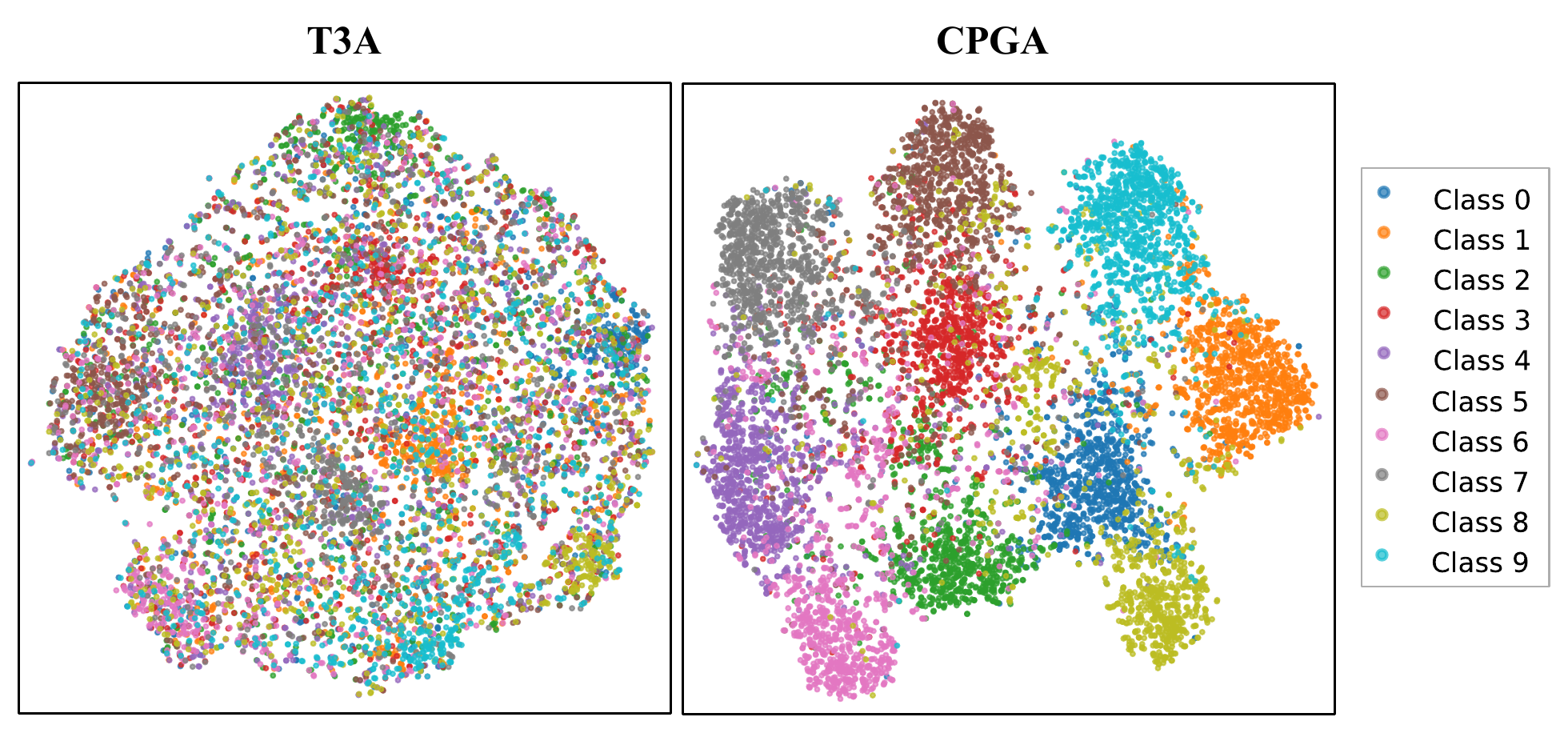}  
    \caption{T-SNE visualization of feature spaces on “Brightness” domain of CIFAR10-C dataset for T3A and our CPGA component.}
    \label{fig:tsne}
\end{figure}

We perform a feature distribution analysis via t-SNE visualization \cite{tsne} on the CIFAR10-C dataset to validate the effectiveness of our proposed CPGA component. Specifically, we replace the CPGA module in BP-TTA with the classifier adjustment module proposed in T3A \cite{T3A}, which is also a prototype-based method. For a fair comparison, all other experimental settings are kept identical. Fig. \ref{fig:tsne} presents the visualization results for the “Brightness” domain within the test stream. As illustrated in the figure, when using T3A, the extracted features from different classes are heavily intermixed, with substantial overlap across categories, making it difficult to form well-separated clusters. In contrast, the features obtained with our CPGA exhibit significantly improved separability, forming more compact and well-defined clusters for each class. These results suggest that CPGA promotes tighter intra-class cohesion and clearer inter-class boundaries within the local feature space by using $\mathcal{L}_{\text{proto}}$.

\subsubsection{Robustness to different Dirichlet concentration parameters}
We vary the value of Dirichlet concentration parameter $\delta$ on three corrupted datasets in Table~\ref{table:delta}. In this setting, the parameter $\delta$ controls the temporal correlation of class labels in the test data stream, where a smaller $\delta$ produces a stronger temporal dependency and leads to a higher degree of class imbalance. As we can see, as $\delta$ decreases from 1e-1 to 1e-6, the average error rates of BP-TTA increase 3.68\%, 2.68\%, and 1.39\% on CIFAR10-C, CIFAR100-C, and ImageNet-C, respectively. Despite the increasingly challenging imbalance conditions induced by smaller $\delta$, the performance degradation of BP-TTA remains relatively limited. These results indicate that the proposed method can maintain stable performance under varying levels of temporal class imbalance, demonstrating its robustness to different Dirichlet concentration settings in the test stream.

\begin{table}[t]
  \centering
  \caption{Average classification error (\%) on CIFAR10-C, CIFAR100-C, and ImageNet-C for different Dirichlet concentration parameters $\delta$}
  \begin{tabular}{c|cccccc}
      \toprule
      \textbf{$\delta$} & \textbf{1e-1} & \textbf{1e-2} & \textbf{1e-3} & \textbf{1e-4} & \textbf{1e-5} & \textbf{1e-6} \\
      \midrule
      CIFAR10-C   & 21.80 & 23.47 & 24.18 & 24.78 & 24.89 & 25.48 \\
      CIFAR100-C  & 35.16 & 36.64 & 37.37 & 37.18 & 37.28 & 37.84 \\
      ImageNet-C  & 66.01 & 66.54 & 66.74 & 66.77 & 67.27 & 67.40 \\
      \bottomrule
  \end{tabular}
  \label{table:delta}
\end{table}

\subsubsection{Robustness to the distribution changing order} 
\begin{figure}[t]
    \includegraphics[height=5cm]{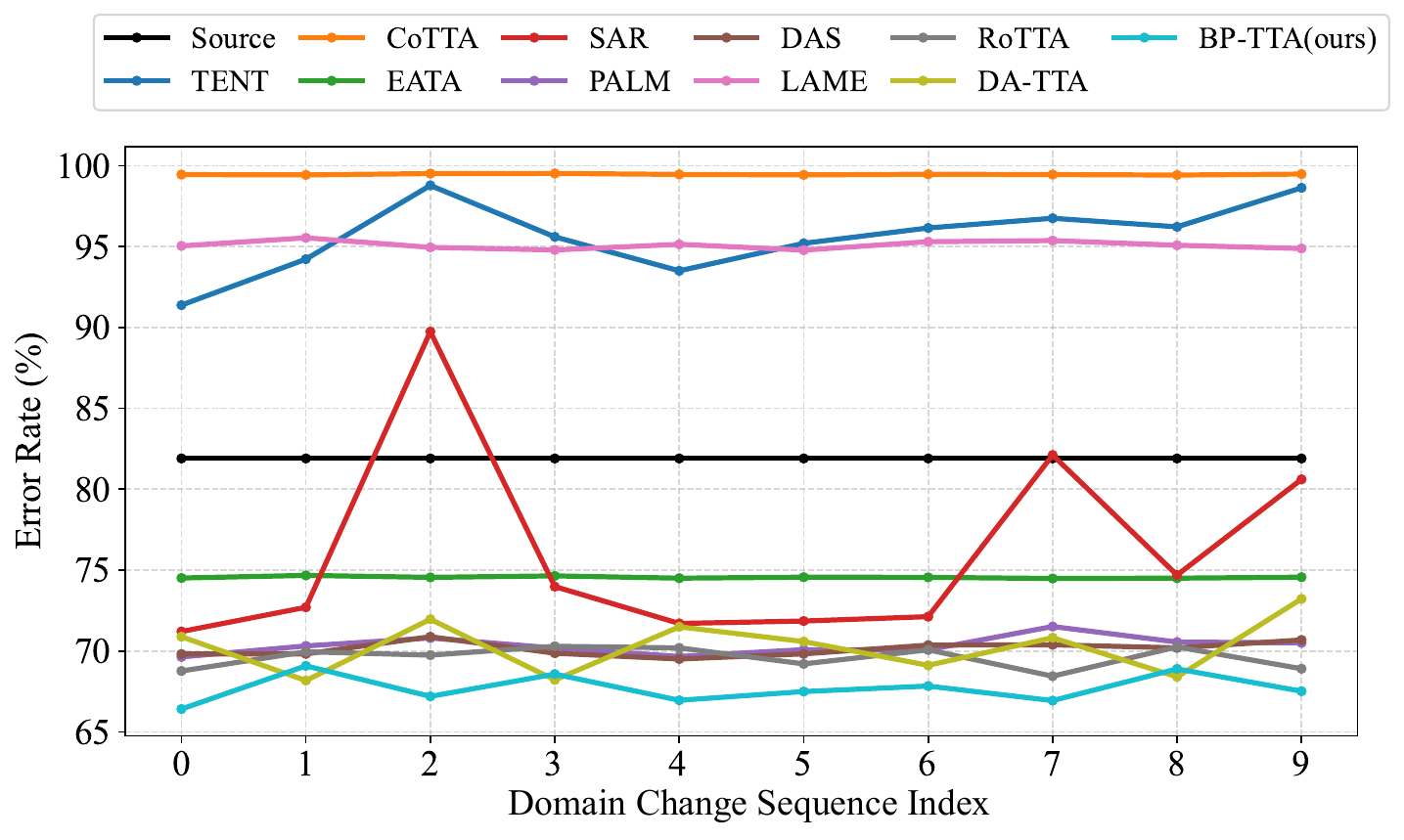}
    \caption{Error rates of different TTA methods on ImageNet-C across 10 distinct predefined sequences of distribution shifts. The x-axis denotes the index of each sequence from 1 to 10.}
    \label{fig:orders}
\end{figure}

To verify that the effectiveness of the proposed method is not dependent on specific orders of data distribution shifts, we conducted experiments with ten distinct sequences of distribution changes on three datasets under the PTTA setting. We report the average classification error of ten different domain shift sequences on three datasets in Table~\ref{tab:order}. Specifically, Fig. \ref{fig:orders} demonstrates the results on ImageNet-C. From the figure, we observe that DA-TTA can achieve competitive performance and even obtain the best results on several specific sequences. However, its performance varies significantly across different sequences, indicating that the method is highly sensitive to the order in which distribution shifts occur and may suffer from large performance fluctuations. In contrast, BP-TTA maintains consistently strong performance across all sequences, demonstrating stronger robustness to sequence ordering and achieving the best average performance. 

\begin{table}[t]
    \centering
    \caption[b]{Average classification error (\%) on CIFAR10-C, CIFAR100-C, and ImageNet-C under 10 different domain shift sequences}
    \begin{tabular}{l|ccc}
        \toprule[1.2pt] 
        Method & CIFAR10-C & CIFAR100-C & ImageNet-C  \\
        \midrule
        Source & 43.52 & 46.45 & 81.90 \\
        TENT \cite{tent} & 85.30 & 97.47 & 95.64 \\
        CoTTA \cite{cotta} & 81.46 & 81.40 & 99.46 \\
        EATA \cite{eata} & 75.23 & 79.75 & 74.55 \\
        SAR \cite{sar} & 75.71 & 91.34 & 76.23 \\
        PALM \cite{palm} & 77.30 & 88.60 & 70.37 \\
        SURGEON \cite{surgeon} & 81.92 & 88.73 & 70.13 \\
        LAME \cite{lame} & 28.54 & 75.93 & 95.09 \\
        RoTTA \cite{rotta} & \underline{26.07} & \underline{39.26} & \underline{69.58} \\
        DA-TTA \cite{da-tta} & 27.03 & 40.38 & 70.28 \\
        \midrule
        BP-TTA & \bf 24.26 & \bf 37.27 & \bf 67.69 \\ 
        \bottomrule[1.2pt]
    \end{tabular}
    \label{tab:order}
\end{table}

\subsection{Comparison with model variation}
\label{sec:models}
To validate the adaptive applicability of our proposed method across various models, we conduct additional experiments. In the CIFAR-C benchmark, we use ResNext-29 \cite{xie2017aggregated}, WideResNet-28 \cite{zagoruyko2016wide}, WideResNet-40 \cite{zagoruyko2016wide}, and PreActResNet-18 \cite{he2016identity} model from \cite{croce2robustbench}. For simplicity, we refer to these models as RNXT29, WRN28, WRN40, and PARN18, respectively. However, the CIFAR100-C pre-trained model parameters for WRN28 is not available in \cite{croce2robustbench}, so we exclude WRN28 from the CIFAR100-C experiment. In the ImageNet-C dataset, we use ResNet-50 \cite{he2016deep}, WideResNet-50 \cite{zagoruyko2016wide}, and ResNeXt-50 \cite{xie2017aggregated} provided by the official torchvision library. To simplify, we refer to these models as RN50, WRN50, and RNXT50, respectively.

\subsubsection{Comparison on CIFAR10-C}
Table~\ref{tab:cifar10_model} presents the average classification errors on CIFAR10-C under the PTTA setting with different models. Our proposed BP-TTA consistently outperforms all baseline methods across all models, achieving the lowest error rates of 21.80\%, 15.76\%, 15.87\%, and 19.78\% on WRN28, RNXT29, WRN40, and PARN18, respectively. Notably, although LAME achieves second-best performance on RNXT29 and WRN40, it suffers from significant performance degradation on PARN18, resulting in an error rate of 79.10\%. In contrast, BP-TTA maintains stable adaptation capabilities across all model architectures and achieves the best overall average performance.

\begin{table}[t]
    \centering
    \caption[b]{Average classification error (\%) on CIFAR10-C for different model architectures under the PTTA setting}
    \begin{tabular}{l|cccc}
        \toprule[1.2pt] 
        Method & WRN28 & RNXT29 & WRN40 & PARN18  \\
        \midrule
        Source & 43.52 & 17.98 & 18.27 & 26.47 \\
        TENT \cite{tent} & 84.35 & 88.76 & 85.91 & 82.11 \\
        CoTTA \cite{cotta} & 81.14 & 77.28 & 76.53 & 75.47 \\
        EATA \cite{eata} & 75.21 & 74.54 & 73.52 & 67.85 \\
        SAR \cite{sar} & 75.83 & 78.45 & 74.34 & 69.49 \\
        PALM \cite{palm} & 77.85 & 88.66 & 87.99 & 86.14 \\
        SURGEON \cite{surgeon} & 82.18 & 87.18 & 81.80 & 76.34 \\
        LAME \cite{lame} & 28.53 & \underline{15.87} & \underline{16.89} & 79.10 \\
        RoTTA \cite{rotta} & \underline{24.88} & 18.09 & 19.35 & \underline{23.64} \\
        DA-TTA \cite{da-tta} & 26.59 & 20.48 & 17.05 & 42.08 \\
        \midrule
        BP-TTA (ours) & \bf 21.80 & \bf 15.76 & \bf 15.87 & \bf 19.78 \\ 
        \bottomrule[1.2pt]
    \end{tabular}
    \label{tab:cifar10_model}
\end{table}

\subsubsection{Comparison on CIFAR100-C}
We also conduct experiments on CIFAR100-C with different models, and the results are summarized in Table~\ref{tab:cifar100_model}. Our method demonstrates an average mean error of 36.64\%, 39.15\%, and 49.32\% in each model framework. Although RoTTA achieves the best performance on PARN18, our method still attains highly competitive results with only a marginal increase of 0.22\%.

\begin{table}[t]
    \centering
    \caption[b]{Average classification error (\%) on CIFAR100-C for different model architectures under the PTTA setting}
    \begin{tabular}{l|ccc}
        \toprule[1.2pt] 
        Method & RNXT29 & WRN40 & PARN18  \\
        \midrule
        Source & 46.45 & 46.75 & 56.87 \\
        TENT \cite{tent} & 96.99 & 95.11 & 97.52 \\
        CoTTA \cite{cotta} & 81.29 & 81.74 & 85.39 \\
        EATA \cite{eata} & 79.68 & 80.76 & 82.95 \\
        SAR \cite{sar} & 92.39 & 89.87 & 91.56 \\
        PALM \cite{palm} & 91.73 & 90.69 & 98.20 \\
        SURGEON \cite{surgeon} & 89.10 & 82.49 & 89.95 \\
        LAME \cite{lame} & 75.80 & 79.39 & 51.98 \\
        RoTTA \cite{rotta} & \underline{38.48} & 41.42 & \bf 49.10 \\
        DA-TTA \cite{da-tta} & 43.69 & \underline{41.34} & 93.82 \\
        \midrule
        BP-TTA (ours) & \bf 36.64 & \bf 39.15 &  \underline{49.32} \\ 
        \bottomrule[1.2pt]
    \end{tabular}
    \label{tab:cifar100_model}
\end{table}

\begin{table}[t]
    \centering
    \caption[b]{Average classification error (\%) on ImageNet-C for different model architectures under the PTTA setting}
    \begin{tabular}{l|ccc}
        \toprule[1.2pt] 
        Method & RN50 & RNXT50 & WRN50  \\
        \midrule
        Source & 81.90 & 78.47 & 78.55 \\
        TENT \cite{tent} & 93.14 & 92.91 & 93.73 \\
        CoTTA \cite{cotta} & 99.40 & 99.77 & 99.49 \\
        EATA \cite{eata} & 74.42 & 73.49 & 75.16 \\
        SAR \cite{sar} & 70.97 & 68.33 & 70.28 \\
        PALM \cite{palm} & 69.82 & 67.70 & 71.52 \\
        SURGEON \cite{surgeon} & 70.57 & 76.05 & 79.68 \\
        LAME \cite{lame} & 94.98 & 91.34 & 92.73 \\
        RoTTA \cite{rotta} & 68.64 & \underline{64.85} & \underline{63.08} \\
        DA-TTA \cite{da-tta} & \underline{68.36} & 66.10 & 65.75 \\
        \midrule
        BP-TTA (ours) & \bf 66.77 & \bf 64.53 & \bf 62.16 \\ 
        \bottomrule[1.2pt]
    \end{tabular}
    \label{tab:imagenet_model}
\end{table}

\subsubsection{Comparison on ImageNet-C}
We further evaluate our method on ImageNet-C with different models, and the results are presented in Table~\ref{tab:imagenet_model}. As can be seen, TENT and CoTTA suffer from severe model collapse on all three models, resulting in extremely high error rates. Our proposed BP-TTA demonstrates outstanding performance across all variations, achieving the lowest average error rates of 66.77\%, 64.53\%, and 62.16\% on RN50, WRN50, and RNXT50, respectively.

\section{Conclusion}
In this paper, we propose Balanced and Prototype-Guided Test-Time Adaptation (BP-TTA), a method designed for dynamic scenarios involving continual domain shifts and class-imbalanced data streams. Specifically, we introduce a Batch-Balanced Sampling (BBS) component that reconstructs the incoming test data stream by selecting samples from majority classes while supplementing minority classes, enabling more reliable adaptation. Furthermore, we proposed a Category Prototype-Guided Adaptation (CPGA) component that maintains dynamic class prototypes to provide stable and discriminative class-level guidance under continual domain shifts. Extensive experiments demonstrate that our method consistently outperforms existing methods and achieves state-of-the-art performance.



\bibliography{ref}
\bibliographystyle{IEEEtran}


 




\vfill

\end{document}